\documentclass[letterpaper, 10 pt, conference]{ieeeconf}  % Comment this line out if you need a4paper

\IEEEoverridecommandlockouts                              % This command is only needed if 
\usepackage{graphicx} % for pdf, bitmapped graphics files
\usepackage{epsfig} % for postscript graphics files
\usepackage{mathptmx} % assumes new font selection scheme installed
\usepackage{times} % assumes new font selection scheme installed
\usepackage{amsmath,amssymb,amsfonts}
\usepackage{cite}
\usepackage{pgfplots}
\pgfplotsset{compat=1.7}
\usepackage{pgfplotstable}
\usepackage{tikz,pgfplots}
\usepackage{physics} 
\usepackage{siunitx} 
\usepackage{enumerate} 
\usepackage{tikz,pgfplots}
\usepackage{verbatim}
\usepackage{color}
\usepackage{mathtools}
\usepackage{breqn}
\usepackage{subcaption}
\usepackage{array}
\usepackage{amsmath}
\usepackage{makecell}
\usepackage{soul}  % we can delete this package later
\usepackage{comment}
\usepackage{textcomp}
\usepackage{subcaption}
\usepackage{algorithm}
\usepackage{algpseudocode}
\usepackage[labelsep=period]{caption}

\usepackage{tabularx}
\usepackage[normalem]{ulem}

\usepackage[]{graphicx}
\usepackage{balance}

\usepackage{kotex} % if you want to use korean
\usepackage{lipsum}

\newcommand{\ks}[1]{\textcolor{black}{#1}} % Kyungseo, blue
\newcommand{\jh}[1]{\textcolor{black}{#1}} % Junha, magenta
\newcommand{\rev}[1]{\textcolor{black}{#1}} % Junha, revision

\title{\LARGE \bf
\rev{Tactile-Proprioceptive Sensor Fusion for Contact Wrench Estimation in Whole-Body Physical Human-Robot Interaction}
}

\author{Junha Min, Junghyeon Ma, Jiwung Kwon, Sunggyu Bae, Joohyung Kim, and $^{*}$Kyungseo Park
\thanks{J. Min, Junghyeon Ma, Jiwung Kwon, Sunggyu Bae and K. Park are with the Department of Robotics and Mechatronics Engineering, DGIST (Daegu Gyeongbuk Institute of Science and Technology), Daegu 42988, Republic of Korea}
\thanks{J. Kim is with the Kinetic Intelligent Machine Lab (KIMLAB), University of Illinois Urbana-Champaign, Champaign, Illinois 61801, USA}
\thanks{$^{*}$Corresponding authors: kspark@dgist.ac.kr}}

\begin{document}
    \maketitle
    \thispagestyle{empty}
    \pagestyle{empty}
    
    % Abstract
    \begin{abstract}
    % \lipsum[1-2]
    % Direct physical guidance is a natural means of teaching and interacting with robots, and robotic skins make a key contribution by enabling sensitive contact sensing and localization. This paper presents a tactile–proprioceptive sensor fusion framework for kinesthetic teaching. Tactile cues from pneumatic skin pads serve as contact indicators that bypass the ambiguity between frictional residues and applied external forces, enabling highly sensitive contact detection without explicit friction identification. We fuse these cues with motor-current–based proprioception to reconstruct multi-axis contact forces on the robot surface. To maintain accuracy during motion, we employ a load-dependent friction model and introduce a temporal convolutional network (TCN) to handle stick–slip transitions, reducing uncertainty at contact onset and yielding smooth, responsive guidance. We validate the approach on a skin-integrated robot arm: (i) multi-axis forces are reconstructed in stationary contacts, and (ii) simultaneous force estimation and kinesthetic teaching are demonstrated. Results indicate improved sensitivity and responsiveness across diverse contact conditions compared with tactile-only and proprioception-only baselines, supporting tactile–proprioceptive fusion as a reliable pathway to safe, intuitive physical human–robot interaction.
     Direct physical guidance is a natural means of teaching and interacting with robots, and robotic skins make a key contribution by enabling sensitive contact sensing and localization. This paper presents a tactile–proprioceptive sensor fusion framework for \rev{natural physical human-robot interaction.} Tactile cues from pneumatic skin pads serve as contact indicators that bypass the ambiguity between frictional residues and applied external forces, enabling highly sensitive contact detection without explicit friction identification. We fuse these cues with motor-current–based proprioception to reconstruct multi-axis contact forces on the robot surface. To maintain accuracy during motion, we \rev{employ a temporal convolutional network (TCN) to mitigate friction hysteresis during} stick–slip transitions, reducing uncertainty at contact onset and yielding smooth, responsive guidance. We validate the approach on a skin-integrated robot arm: (i) multi-axis forces are reconstructed in stationary contacts, and (ii) simultaneous force estimation and kinesthetic teaching are demonstrated. Results indicate improved sensitivity and responsiveness across diverse contact conditions compared with tactile-only and proprioceptive-only baselines, supporting tactile–proprioceptive fusion as a reliable pathway to safe, intuitive physical human–robot interaction.
\end{abstract}

    % Main Text
    \section{INTRODUCTION} %% 6 paragraph - each paragraph 5~10 line
 In human–robot interaction, teaching robots through direct physical guidance is regarded as one of the most intuitive approaches. To enable this, reliable perception of physical contact is essential, and numerous approaches to external force estimation have been proposed \cite{paper1_survey1, paper2_survey2}.

 Force/torque (F/T) sensors and joint torque sensors have been used for direct measurement of external forces. While these devices provide accurate measurements, they present several drawbacks, including high cost, susceptibility to damage from external impacts, and reductions in the robot's structural stiffness \cite{paper3_ft_sensor, paper4_torque_sensor1, paper5_torque_sensor2, paper6_proprioceptive1}. As an alternative, many studies estimate joint torques and external forces from motor currents in conjunction with a dynamic model \cite{paper7_proprioceptive2}. This proprioceptive approach improves durability and accessibility by avoiding additional sensors. However, significant errors arise from motor friction and hysteresis, especially in stationary and quasi-static regimes \cite{paper8_friction}. 
 
 To mitigate this issue, larger thresholds are typically introduced to distinguish external forces from model errors. This method is intuitive and simple but effectively introduces a dead band, which degrades the robots' contact responsiveness. To move beyond ad-hoc thresholding, several studies have leveraged signal-processing methods, such as Kalman filtering, to compensate for friction-induced hysteresis \cite{paper9_KF1, paper10_KF2}. More recently, data-driven approaches using artificial neural networks have been explored to learn the nonlinear hysteresis patterns and reduce residual errors \cite{paper11_MD_HRDL,paper12_DUNN,paper13_PINN}.

 In parallel, whole-body robot skins have been considered a promising solution because they enable high-sensitivity contact detection and localization across the robot's entire surface \cite{paper14_comprehensive, paper15_EIT_kspark, paper16_camera_skin}. Such systems should provide not only tactile sensing but also passive safety via mechanical compliance. For instance, pneumatic robot skins provide passive safety through a compliant, impact-absorbing envelope, while internal pressure changes under external loads enable tactile sensing \cite{paper17_soft1}. Moreover, combining mechanical compliance with tactile feedback has enabled safe whole-body manipulation using simple control strategies \cite{paper18_ponyo}. 3D printing has also been leveraged to implement low-cost and easy-to-build skins, and associated interaction frameworks have been demonstrated \cite{paper19_pneumatic_kspark}. Nonetheless, individual skin pads provide only the magnitude of contact force, usually the normal component. 
 
 To address this limitation, sensor-fusion methods that combine joint-torque measurements with robot-skin tactile sensing have been proposed \cite{paper20_icub, paper21_virtual_sensor, paper22_tactile_proprioceptive}; however, factors such as cost and scalability have not been adequately addressed.

 In this paper, we propose a hybrid tactile-proprioceptive approach that combines motor-current measurements with pneumatic robot skins for natural interaction. The main contributions are as follows:
 
 \begin{enumerate}
    \item Using tactile signals as contact cues, we disambiguate static-friction residuals from true external forces, enabling sensitive force perception and agile responses.

    \item By fusing tactile cues with proprioceptive sensing, we reconstruct physical interactions as a multi-axis force vector defined on the robot's surface.

    \item We train a temporal convolutional network (TCN) that models friction from quasi-static residuals and compensates it online. By integrating with the above components, it substantially reduces the dead band and enables a natural, intuitive kinesthetic teaching framework.
 \end{enumerate}

 The remainder of this paper is organized as follows: Section II reviews robot dynamics and external-force estimation. Section III describes the hardware setup, including manipulator and robot skin modules. Section IV presents the proposed sensor-fusion strategy and the TCN-based residual compensator. Section V presents the experiments and demonstration. Finally, Section VI concludes with limitations and direction for future work.

    \section{SYSTEM DYNAMICS MODELING}
    \subsection{Robot Dynamics}
    %n-자유도(DOF) 강체 로봇의 동역학 방정식은 일반적으로 라그랑주-오일러 방정식을 통해 다음과 같이 표현된다. 
    For an $n$-degree-of-freedom (DoF) rigid robot, the joint-space dynamics are written as
    \begin{equation}
        \tau_{dyn} = M(q)\ddot{q} + C(q, \dot{q})\dot{q} + G(q) \quad
    \end{equation}
    
    \rev{where $q, \dot q, \ddot q \in \mathbb{R}^n$ denote the joint position, velocity, and acceleration vectors. The terms $M(q) \in \mathbb{R}^{n \times n}$, $C(q,\dot q) \in \mathbb{R}^{n \times n}$, and $G(q) \in \mathbb{R}^n$ represent the inertia matrix, Coriolis and centrifugal matrix, and gravity vector, respectively. $\tau_{dyn}$ is the inverse-dynamics torque, so the joint-side actuator torque $\tau$ is written as}
    \begin{equation}
        \tau = \tau_{ext} + \tau_{dyn} + \tau_{fric} + \tau_{error} \quad
        \label{eq:2}
    \end{equation}
    where $\tau_{\mathrm{ext}}$ the torque induced by external contacts, $\tau_{\mathrm{fric}}$ the internal friction torque, and $\tau_{\mathrm{error}}$ aggregates modeling error and measurement noise. 
    
    % 여기서 q,q˙,q¨∈Rn는 각각 관절의 위치, 속도, 가속도 벡터를 의미한다. $M(q) \in \mathbb{R}^{n \times n}$는 관성 행렬, $C(q, \dot{q}) \in \mathbb{R}^{n \times n}$는 코리올리 및 원심력 행렬, G(q)∈Rn는 중력 보상 벡터, 그리고 τdyn∈Rn는 순수 동역학에 의한 관절 토크를 나타낸다. 로봇의 관절 구동기에서 측정되는 전체 토크(τ)는 동역학 토크(τdyn), 마찰 토크(τfric), 그리고 외부 환경과의 접촉으로 인한 외력 토크(τext)의 합으로 구성된다. 따라서 외력 토크는 전체 측정 토크에서 동역학 및 마찰 모델 값을 감산하여 추정할 수 있으며, 이 과정에서 발생하는 모델링 불확실성과 노이즈는 오차 항(τerror)으로 정의된다. 
        
    %작업 공간에서의 외력(Fext)과 관절 공간에서의 외력 토크(τext)는 자코비안 행렬(J)을 통해 다음과 같은 관계를 가진다. 
    The task-space external wrench $F_{ext}\in \mathbb{R}^{6}$ and the corresponding joint-space external torque $\tau_{ext}$ are related via the Jacobian $J(q)$:
    \begin{equation}
        \tau_{ext} = J(q)^T F_{ext}.
        \label{eq:jacobi}
    \end{equation}
    %%%%%%%%%%
    %  패드별로 dh파라미터가 어떤식으로 정의됐는지, 이에 따른 자코비안 계산 방식은 어떤지.
    The geometric Jacobian $J(q)$ in (3), which relates joint velocities to the contact point velocity, is computed from the joint origins $P_i$, rotation axes $z_i$, and the contact point $P_{\mathrm{contact}}$, all expressed in the base frame $\{0\}$. The contact point $P_{\mathrm{contact}}$ can be defined by introducing an additional frame estimated from tactile sensor measurements. For a revolute joint $i$, the corresponding Jacobian column $[J_{v,i}^T, J_{w,i}^T]^T$ is given by the standard formulation.

    \begin{equation}
        J_{v,i} = z_i \times (P_{\mathrm{contact}} - P_i), \quad J_{w,i} = z_i
    \end{equation}
    where $J_{v,i}$ and $J_{w,i}$ represent the linear and angular velocity contributions. The resulting Jacobian is then transformed to the local contact frame using the rotation matrix  $R_{\mathrm{contact}}$.

    \subsection{Friction Models}
    In contrast to the dynamics model, friction is highly sensitive to various physical properties, making its modeling more challenging. Accurate identification of $\tau_{fric}$ is thus essential for the precise estimation of the external torque $\tau_{ext}$. 
    
    A commonly used static friction model combines Coulomb friction $\tau_{c}$, viscous $\tau_{v}$, and Stribeck effects:
    \begin{equation}
      \tau_{fric}(\dot q)\;=\;\Big[\tau_c + (\tau_s - \tau_c)\exp\!\Big(-\Big|\tfrac{\dot q}{\dot q_s}\Big|^{\delta}\Big)\Big]\operatorname{sgn}(\dot q)\;+\; b\,\dot q
      \label{eq:static_fric}
    \end{equation}
    where $\tau_s$ is the static level, $b>0$ the viscous coefficient (equivalently $\tau_v(\dot q)\triangleq b\,\dot q$), $\dot q_s>0$ the Stribeck velocity scale, and $\delta>0$ a shape parameter.
    % 위 식에서 Fc, Fs, Fv는 각각 coulomb 마찰, staic 마찰, viscous 마찰 계수를 나타낸다.
    
    \rev{For high-reduction actuators (ratio $\approx$ 500:1) used in this study, the internal contact forces in the transmission scale significantly with the applied load, directly influencing friction dynamics. To account for this, the constant friction coefficients in \eqref{eq:static_fric} are modified to include load-dependent functions 
    \cite{paper23_load_friction}:
    \begin{equation}
        \tau_{c,load} = \tau_c + \alpha_c |G(q)|,\quad b_{load} = b + \alpha_v |G(q)| 
    \label{eq:load_fric}
    \end{equation}
    where $\tau_{c,load}$ and $b_{load}$ denote the load-dependent Coulomb and viscous friction coefficients, and $\alpha_c,\alpha_v\ge 0$ are scaling constants characterizing load dependence. Here we use $\lvert G(q)\rvert$ as a proxy for load; this approximation is acceptable because the robot links are the primary source of load, whereas pHRI interaction forces were assumed to be gentle in typical physical guidance. }
    
    \rev{While the aforementioned friction models describe motor friction reasonably well, they do not capture the complex dynamics, including stick-slip motion and nonlocal memory hysteresis. Although advanced models like LuGre \cite{paper24_LuGre} or Generalized Maxwell-Slip (GMS) \cite{paper25_GMS} address these phenomena, they require identifying a large number of parameters and are sensitive to environmental changes.
    }
    
    \rev{To this end, we avoid complex physical modeling and employ a friction dynamics model based solely on \eqref{eq:static_fric} and \eqref{eq:load_fric}. Instead, we leverage tactile information to compensate for frictional hysteresis, effectively bypassing the intricate parameter identification process as detailed in the following sections.}

    \section{Skin-integrated Robot Arm}
    % \begin{figure}[t!]
    %     \centering
    %     \includegraphics[width=3.3in]{Figure/figure_one_col.png}
    %     \caption{Rendered image of the Skin-integrated robot arm.Sensing electronics.}
    %     \label{figure label 05}
    % \end{figure}
    
    \subsection{Manipulator}

        \begin{figure*}[htbp]
            \centering
            \includegraphics[width=6.6in]{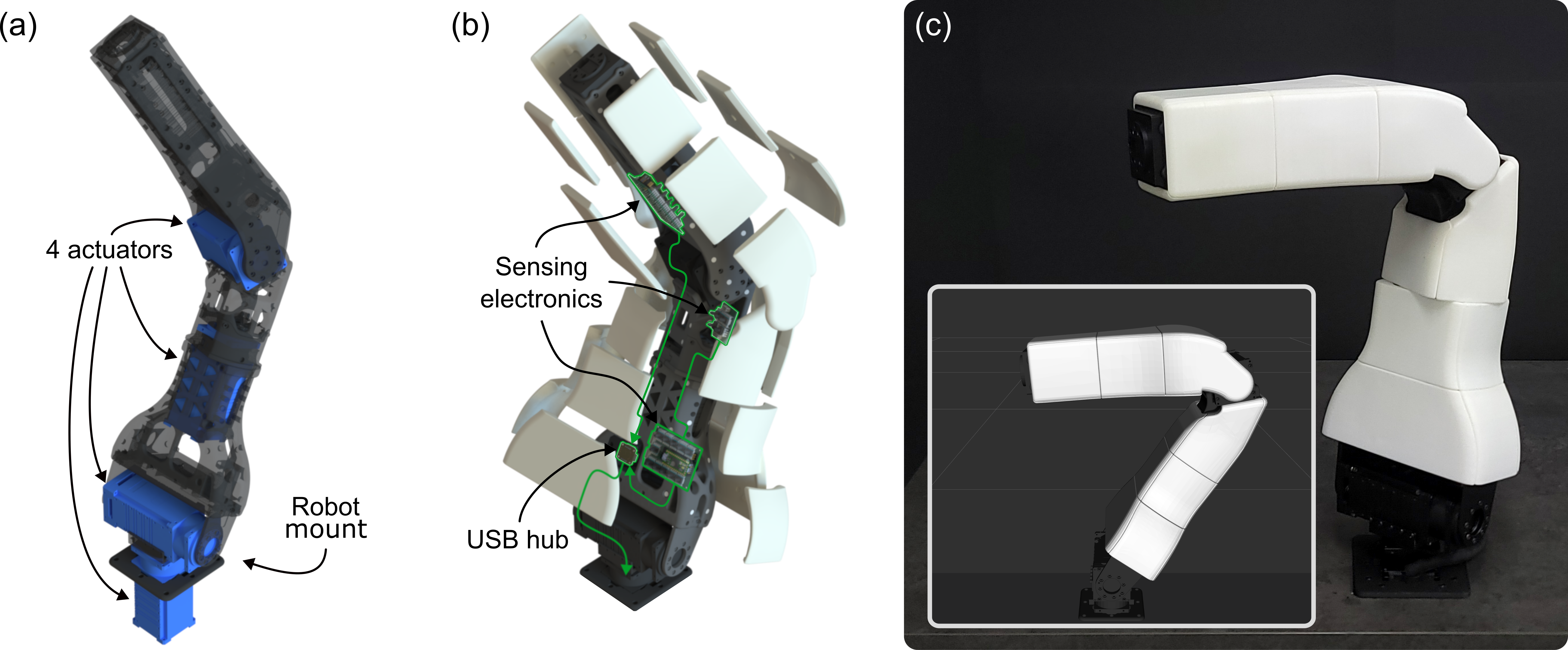}
            \caption{Skin-integrated robot arm. (a) Kinematic structure and actuator layout of the manipulator. (b) Pneumatic skin design and sensing electronics connections. (c) Photograph of the skin-integrated robot arm and its RViz visualization.}
            \label{figure label 01}
            \vspace{-6pt}
        \end{figure*}  
        
        We conducted our study using a serial manipulator integrated with pneumatic robot skin. The platform is a conventional elbow-type manipulator \cite{paper26_papras}. To enable more natural motion, we added a new actuator on the proximal link whose rotational axis is coaxial with the link’s longitudinal axis. The wrist joint was removed to simplify system identification and emphasize whole-body interaction. As a result, our setup has four DoFs at the forearm, as illustrated in {Fig. 1(a)}.

        Each joint is actuated by a servo motor with a cycloidal gear reduction of approximately 500:1 (DYNAMIXEL-P PH54 series, Robotis). The motors use a 24 V supply power and communicate with peers and the host through an RS-485 daisy chain. The communication cycle was nearly 400 Hz, so motor state variables and configuration parameters (e.g., control gains) can be exchanged in real-time.  
        %%%%%% 추가로 할말? %%%%%%

        %앞서 설명하였듯 본 연구에서 사용한 하드웨어 시스템은 우리의 이전 연구에서 제안한 Skin-integrated된 시리얼 매니퓰레이터(Plug and play robotic arm system) [tro], [papras]를 기반으로 하며 기존의 Plug-and-play feature를 유지하였다. 본 연구에서는 동역학 모델 identification 과정을 단순화하기 위하여 4자유도로 구성하였으며, actuactor는 501.9 : 1 cycloid 감속기를 가지는 PH54-200-S500-R, PH54-100-S500-R 모터를 사용하였으며 rs485통신을 통해 모터의 mcu와 400hz의 통신주기를 가진다.

        % \begin{equation}
        %  \frac{\Delta P}{P_0} = \left(1 + \frac{\Delta V}{V_0}\right)^{-1} - 1  
        % \end{equation}

        % Skin-integrated 로봇 디자인
        % Components (motor, link, skin, ...)
        % Plug-and-play feature
        % design consideration (pad dimension, etc.)
        % Fabrication details
        % 이미 예전에 발표한 연구 내용 -> 너무 자세하게 설명할 필요는 없음
    
    \subsection{Pneumatic Robot Skin} 
        \rev{We adopted the pneumatic robot skin framework proposed in our previous work [19] and customized it for the manipulator, as shown in Fig. 1(b).}
        % The pneumatic robot skin was customized for the manipulator, as shown in Fig. 1(b).
        Each skin pad was 3D-printed (FDM-type) from flexible TPE (eLastic, eSUN) and chemically smoothed with tetrahydrofuran (THF) to achieve air-tight surface. Each pad has a barbed port and was connected via silicone tubing to an air-pressure sensor (HSCDANT005PGAA5, Honeywell) placed inside the robot. The sensor outputs 0 -- 3.3 V in proportion to the applied air pressure, and microcontroller board (Teensy 4.0, PJRC) reads those values in real-time.

        The microcontroller handled low-level signal processing (e.g., low-pass filtering, contact thresholding) and data transmission. Data were formatted as arrays and sent to the host's ROS 2 network via micro-ROS with a publishing rate of 1 kHz.

        The pneumatic robot skin was sensitive to small external forces, and its bandwidth exceeded that of motions from typical pHRI scenarios, enabling diverse interactions from basic teaching to gesture-based commanding \cite{paper19_pneumatic_kspark}. However, within a single pad, contact localization was not possible and only the normal force could be measured. Moreover, for the same applied force, the pressure reading varied with contact area and geometry. These limitations impose significant constraints on tactile-only approach and motivate complementary hybrid with proprioception. 
        \vspace{3pt}

        % Each pad has a barbed port so it can be connected to the pressure sensor (모델명, Honeywell) through a urethane tube, and the pressure sensors date are measured by microcontroller (Teensy 4.0). The pcb that includes mcu & pressure sensors is located inside of each link. 공압 센서는 외력 감지에 대한 0.05kPa의 deadband(threshold)를 설정하였으며 raw데이터 고주파 노이즈 제거를 위해  40hz의 cut off frequency를 가지는 1차 iir필터가 적용 되어있다. Signal processing을 거친 데이터 array는 micro ros를 통해 중앙 ROS 시스템으로 1000hz의 주기로 발행된다.
        % Sensor configuration (PCB, teensy, connection, ...)
        % signal processing (LPF, deadband, ...)
        % data transmission (microROS, sampling frequency, ...)

        %tactile sensor는 지속적인 힘, 순간적 이벤트와 같은 복합적인 접촉을 감지할 수 있으며 기계적 compliance를 통한 충격 완화적 특성을 가진다. 주요 목적은 낮은 해상도에서 대략적인 외력 위치를 감지하는 것(We have decided not to pursue precise contact localization performance in terms of spatial resolution.)이다. 로봇의 skin pad의 material은 tpu로, 3d프린터를 이용하여 출력되며 자이로이드 내부 패턴을 가지므로, 동일한 패드상의 임의의 deform은 그 위치와 무관하게 내부 압력 변화를 유도한다. 이때 압력과 부피 변화는 다음과 같은 관계를 가진다.
    % \input{Chapter/04_Sensor fusion}
    \section{SENSOR FUSION}
    In robotic systems, the measured torque $\tau_{meas}$ is typically decomposed into several components, as shown in Equation \eqref{eq:ext_force}. This measured torque is often estimated from the motor current $I$, and a torque constant $K_t$ via Equation \eqref{eq:torque_current}.  
    \begin{equation}
        \tau_{meas} = \tau_{dyn} + \tau_{fric} + \tau_{res}
        \label{eq:ext_force}
    \end{equation}
    \begin{equation}
        \tau_{meas} = K_{t} I.
        \label{eq:torque_current}
    \end{equation}
    Here, the torque residual $\tau_{res}$	in Equation \eqref{eq:ext_force} comprises not only the external torque $\tau_{ext}$ but also inherent modeling errors and complex frictional behavior not captured by nominal model. These unmodeled effects include static friction and stick–slip transitions under static and quasi-static conditions. Such friction components exhibit strong hysteresis dependency, which makes their analytical modeling exceptionally challenging and complicates the accurate estimation of $\tau_{ext}$.

    To address this issue, this paper proposes two methods. First, we use tactile signals to estimate static friction and to infer external forces under static conditions. Second, we introduce a learning-based method that handles uncertainty at the static-to-kinetic transition, enabling more natural motion in pHRI tasks such as kinesthetic teaching.

\subsection{Contact-aware Torque Estimation}
% 이를 해결하기 위하여 우리는 tactile데이터를 활용하여 정지상태에서의 torque residual을 구성하는 static frictoin과 external torque를 분리하는 방법론을 제안한다. q'=0인 상황에서 Static friction은 다음과 같은 관계를 가진다.
   % \begin{equation}
   %      \tau_{staticfric} = \tau_{redisual} - \tau_{ext}
   %  \end{equation}
% 이때 대면적 tactile는 $\tau_{ext}$의 작용 여부를 측정할 수 있으므로, 외력이 작용(접촉) 시점 기준 $\tau_{redisual}$의 변화율인 $\Delta\tau_{redisual}$를 계산할 수 있으며 이를 통해 다음과 같이 static friction와 $\tau_{ext}$는 다음과 같이 계산할 수 있다.
    % \begin{equation}
    %     \hat{\tau}_{staticfric} =
    %         \begin{cases}
    %             \tau_{redisual} - \Delta\tau_{redisual}& \text{if } \dot{q} = 0 
    %             \\0 & \text{otherwise}
    %         \end{cases}
    %     \label{eq:static_friction_def}
    % \end{equation}

    % \begin{equation}
    %     \hat{\tau}_{ext} = \Delta\tau_{redisual} = \Delta\tau_{measure}
    %     \label{eq:static_ext_def}
    % \end{equation}
% 이 정의는 촉각 접촉이 감지된 이후, 외부 토크가 얼마가 가해지든 $\hat{\tau}_{\text{static friction}}$ 값은 접촉 직전의 값($t_{\text{contact}}^{-}$)으로 
% 일정하게 유지됨을 보장한다. 또한 정적 상황에서 $\Delta\tau_{redisual}$와 $\Delta\tau_{measure}$가 같으므로 동역학 모델 없이 ${\tau}_{ext}$를 추정할 수 있다. 

    % method는 개별 방법, methodology는 방법론 (여러 방법을 아우르는 원리 등에 관한 내용)
    
    We propose a method that leverages tactile data to disambiguate static friction from external torque. Under static conditions ($\dot{q}=0$), the torque residual is modeled as the sum of static friction $\tau_{fric, s}$ and external torque $\tau_{ext}$:
       % \begin{equation}
       %      \tau_{redisual} = \tau_{staticfric} + \tau_{ext}
       %  \end{equation}

    \begin{equation}
        \tau_{res} \;=\; \tau_{fric, s} \;+\; \tau_{ext} .
    \end{equation}
    Using tactile signals, we can detect the physical contact and its onset time $t_{on}$, and define the pre-contact baseline $\tau_{res}(t_{on}^{-})$. We then compute the increment of the residual:
    \begin{equation}
        \Delta \tau_{res}(t) \;=\; \tau_{res}(t) \;-\; \tau_{res}(t_{on}^{-}).
    \end{equation}
    Here, we can assume that the pre-contact baseline $\tau_{res}(t_{on}^{-})$ can be attributed exclusively to friction.
    \begin{equation}
        \hat{\tau}_{fric, s} \;=\; \tau_{res}(t_{on}^{-}).
        \label{eq:static_friction_def}
    \end{equation}
    While the robot remains at rest, we assume the static-friction term is constant for small external loads such as gentle physical guidance. Consequently,
    \begin{equation}
        \hat{\tau}_{ext}(t) \;=\; \Delta \tau_{res}(t) \;=\;  \Delta\tau_{meas}.
        \label{eq:static_ext_def}
    \end{equation}
    In this way, we could estimate the torque due to external forces without relying on a dynamics model.
% % % % % % % % % % % fig2% % % % % % % % % % % % % % 
    \begin{figure}[t!]
        \centering
        \includegraphics[width=2.2in]{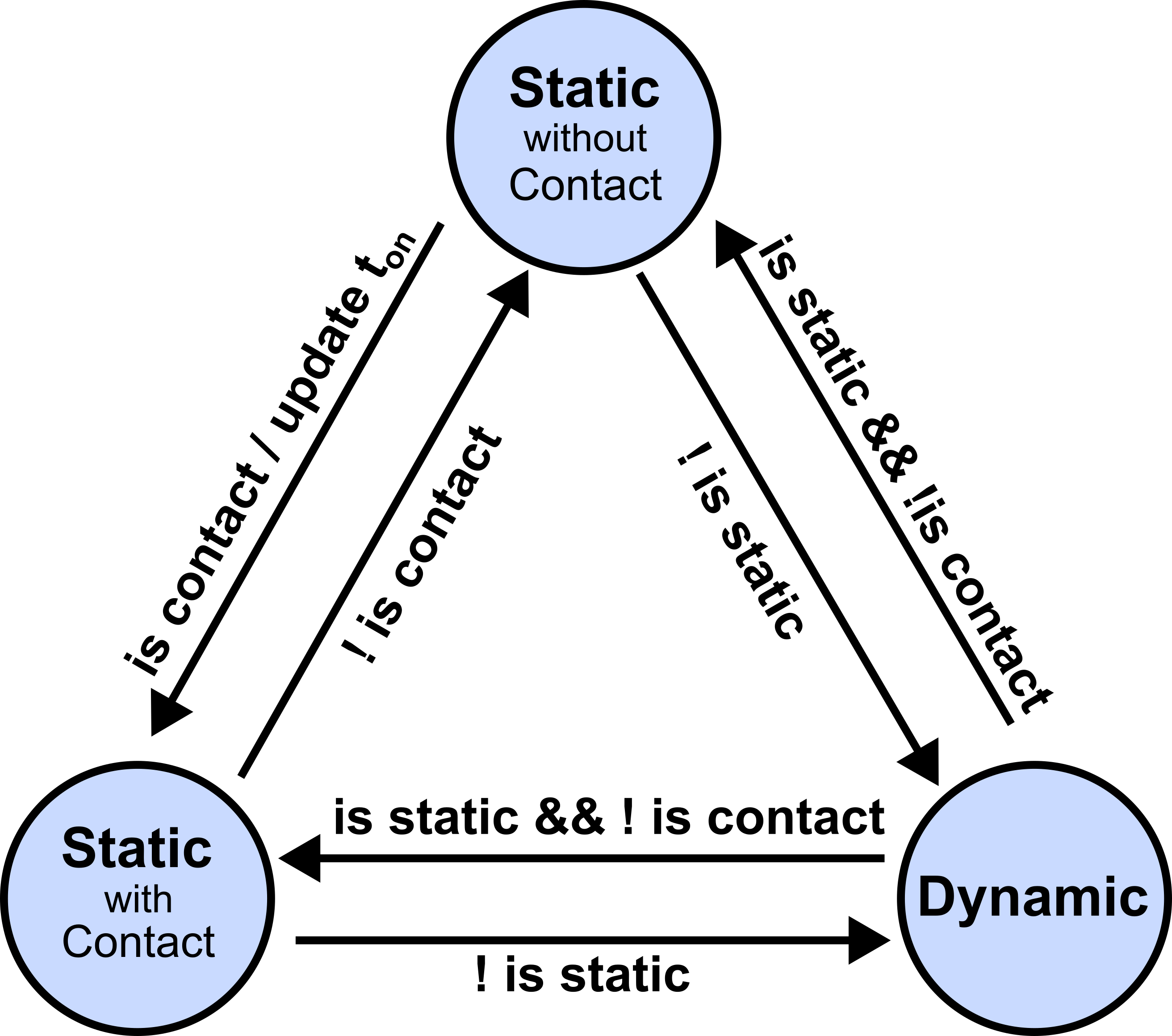}
        \caption{Contact-aware torque estimation algorithm represented as a three-state Finite-State Machine (FSM). Transition conditions: (1) 'is contact' indicates tactile sensor activation, (2) 'is static' indicates stationary robot state, and (3) '$\text{update } t_{on}$' sets the contact onset reference time.}
        \label{FSM}
    \end{figure}
    The proposed algorithm can be formalized as a finite-state machine (FSM), as shown in Fig. 2(b). Here, we consider three states:
    \vspace{6pt}
    
    \noindent {(i) Static (without contact)\\: the robot is stationary and no external contact is detected.}
    \vspace{6pt}

    \noindent {(ii) Static (with contact)\\: a contact is detected while the system remains stationary.}
    \vspace{6pt}
    
    \noindent {(iii) Dynamic\\: the robot is in motion (i.e., physical interaction).}
    \vspace{6pt}

    As contact occurs, transitioning the state from (i) to (ii), these tactile cues are used to estimate the external force via Equation \eqref{eq:static_ext_def}.
    
    This approach improves responsiveness in high-gear-ratio systems, where friction introduces large uncertainty. By canceling static friction at contact onset, we can bypass the large dead-band, which is required by other methods that rely purely on motor current and a dynamics model.

    However, this method is valid only in the static states (i) and (ii). Once motion is initiated by the estimated external force, Equation \eqref{eq:static_ext_def} is no longer valid. 
    % During this phase, the static friction, which continuously  
    \jh{During this phase, the friction transitions from static to dynamic and exhibits trajectory-dependent hysteresis, which cannot be captured by Equation} \eqref{eq:static_friction_def}.
    % during the transition from static to kinetic regimes and exhibits hysteretic characteristics dependent on the prior motor trajectory, cannot be described by Equation %\ks{\eqref{eq:static_friction_def}}
    Consequently, a large discontinuity arises when transitioning from state (ii) to state (iii).

    \subsection{TCN-Driven Seamless Transition Friction Compensator}
    Friction in the static to kinetic transition is hysteretic and \jh{nonlinear}, so it cannot be fully characterized from the instantaneous kinematic states ($q,\dot{q},\ddot{q}$) alone. 
    % states of the art 가 매소드에서 또나오는 느낌 < 차라리 인트로로 좀 보내야겠다
    % Memory-based models (e.g., LSTMs) use kinematic histories to address hysteretic behavior, but they still provide no direct observation of the latent internal friction state \cite{paper27_icra2017, paper11_MD_HRDL}. Moreover, the static-friction level exhibits aleatoric, trial-to-trial variability with high variance. Consequently, such approaches are fundamentally limited.
    To address this challenge, we propose a data-driven compensation strategy based on a Temporal Convolutional Network (TCN). The proposed model utilizes not only the time-series data of the kinematic state $q,\dot{q},\ddot{q}$ but also the estimated static friction $\hat{\tau}_{fric, s}$ as an additional input. 
    \rev{The estimate $\hat{\tau}_{fric, s}$, explicitly captured by the FSM via tactile cues at motion onset, provides the system with the precise initial condition for the moment of static-to-dynamic state transition.}
    % This $\hat{\tau}_{fric, s}$ value, isolated from $\tau_{res}$ via tactile contact cue, is not merely an estimate; this provides the system with the precise initial condition for the moment of static-to-dynamic state transition.
    This is key information to explicitly resolve the aforementioned ambiguity, which could not be addressed using only kinematic data. By referencing this initial friction state, the TCN processes the subsequent kinematic time-series data to effectively infer the complex friction residual during the transition phase, which the conventional friction model \eqref{eq:static_fric} fails to estimate. \jh{The TCN is suited for efficiently learning this short-term, hysteretic data.}
    % To address this challenge, this study proposes a compensator based on a Temporal Convolutional Network (TCN). This network utilizes not only the time-series data of the kinetic state but also the estimated static friction $\hat{\tau}_{staticfric}$ derived from Equation \eqref{eq:static_friction_def} as an additional input. This $\hat{\tau}_{staticfric}$ value, isolated from $\tau_{residual}$ via tactile contact data, is not merely an estimate; it provides the system with the precise initial condition at the moment the dynamic transition begins. This provides the critical information required to explicitly resolve the aforementioned ambiguity, which could not be addressed using only kinetic data. By referencing this initial friction state, the TCN processes the subsequent kinematic time-series data to effectively infer the complex friction residual during the transition phase, which the conventional friction model \eqref{eq:static_fric} fails to estimate.
    % 이러한 문제를 resolve 하기 위하여, 본 연구는 Temporal Convolutional Network(TCN) 기반 보상기를 제안한다. 이 네트워크는 kinetic 상태의 시계열 데이터뿐만 아니라, Method 1의 Equation \eqref{eq:static_friction_def}를 통해 추정된 정지 마찰 $\hat{\tau}_{staticfric}$ 값을 추가 입력으로 활용한다. Tactile sensor의 접촉 정보를 통해 torque residual $\tau_{residual}$ 로부터 분리된 $\hat{\tau}_{staticfric}$ 값은 단순한 추정치가 아니라, 동적 전환이 시작되는 순간의 정확한 initial condition을 시스템에 제공한다. 이는 기존에 kinetic 정보만으로는 해결할 수 없었던 앞서 언급된 ambiguity를 explicitly하게 해결하는 핵심 단서(key value?)가 된다. TCN은 초기 마찰 상태 값을 기준으로, 이후 입력되는 운동학적 시계열 데이터를 처리하여 기존 마찰 모델 \eqref{eq:static_fric}이 추정하지 못하는 복잡한 전환 구간의 friction residual를 효과적으로 추론할 수 있다. 
    \begin{figure}[t!]
        \centering
        \includegraphics[width=3.3in]{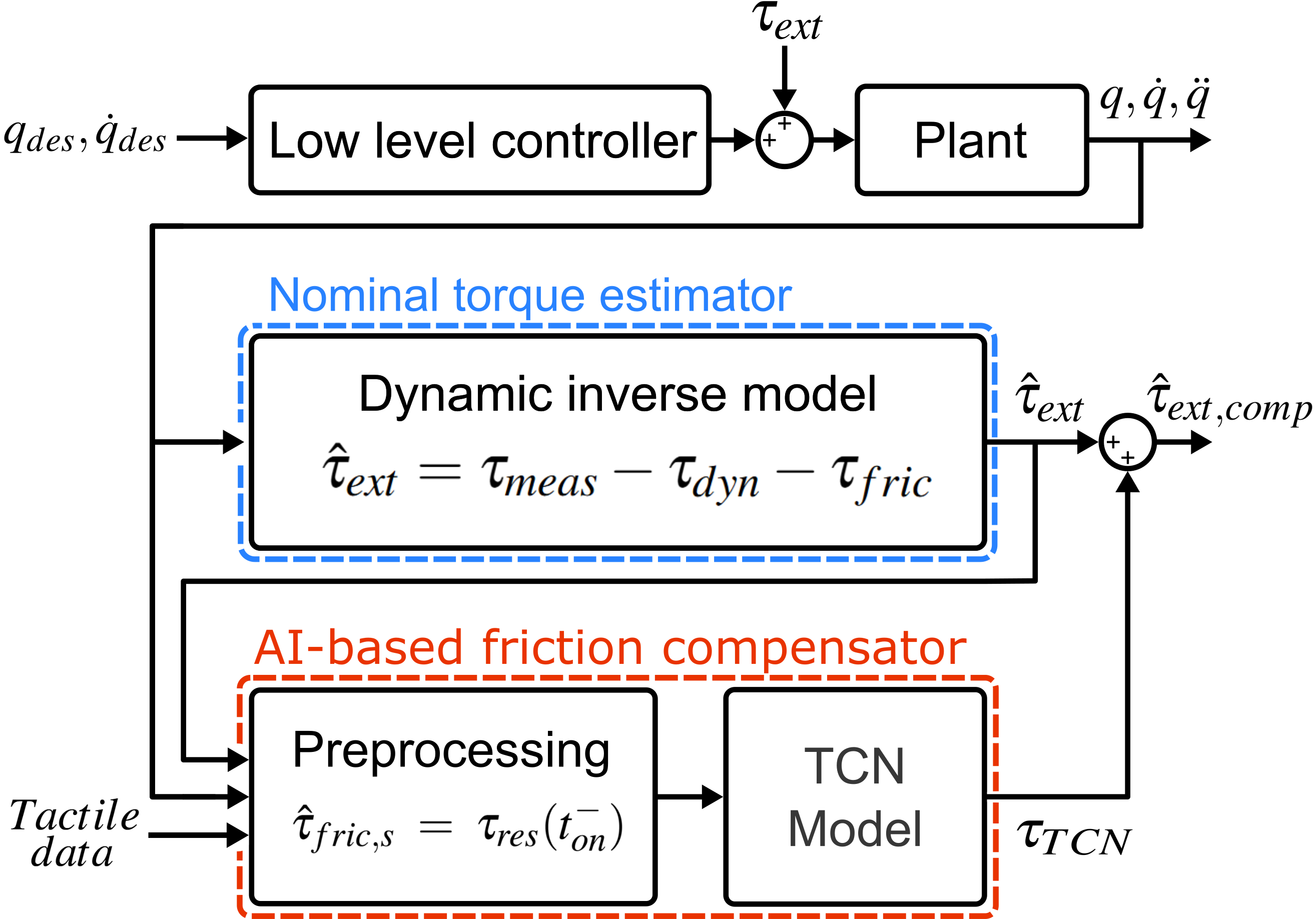}
        \caption{Pipeline of the joint external torque estimation. An AI-based friction compensator utilizes the kinetic state, the output derived from the nominal inverse dynamics model, and tactile sensor data to compensate for static friction and friction within the transition region.}
        \label{figure label 03}
    \end{figure}
    
    \begin{equation}
        \tau_{\text{TCN}} = TCN(q, \dot{q}, \ddot{q}, \hat{\tau}_{\text{fric,s}}), \; \jh{(\hat{\tau}_{\text{fric,s}} = 0 \text{ if } \dot{q} \neq 0).}
        \label{eq:TCN_input}
    \end{equation}
    Constraining the network input to the static-friction estimate $\hat{\tau}_{fric, s}$ is crucial \jh{for inference under forced conditions.} 
    Feeding the non-disambiguated residual $\tau_{res}$ would cause the network to compensate not only friction but also external torque, thereby degrading external-torque estimation. The tactile sensing stage separates friction from external torque, ensuring that the network receives an input invariant to external forces.
    %Constraining the TCN input exclusively to $\hat{\tau}_{staticfric}$ is essential for robust generalization. If the TCN utilized the non-disambiguated $\tau_{res}$ directly, the network would be trained to compensate not only for the friction component but also the external torque component, consequently diminishing the system's external torque estimation capability. In other words, the tactile sensor performs the critical role of separating the friction and external torque components from $\tau_{residual}$, ensuring that the TCN input consists only of terms independent of the external torque. This implies that even if the TCN is trained on data lacking external torque, it can generalize effectively during real-time inference processes where external forces are present.    
    % TCN의 입력 설계를 $\hat{\tau}_{staticfric}$로 한정하는 것은 강건한 일반화(robust generalization) 성능에 필수적이다. 만약 TCN이 분리되지 않은 $\tau_{residual}$ 값을 직접 입력으로 사용할 경우, 네트워크는 마찰 성분뿐만 아니라 외력 성분까지 함께 보상하도록 학습되어, 결과적으로 시스템의 외력 추정 기능을 상실하게 된다. 즉, 촉각 센서는 $\tau_{residual}$로부터 마찰 성분과 외력 성분을 분리하여, TCN의 입력이 외력과 무관한 항으로만 구성되도록 보장하는 핵심 역할을 수행한다. 이는 곧 외력이 포함되지 않은 데이터로 TCN을 학습시키더라도, 실제 외력이 존재하는 실시간 추론 과정에서도 일반화가 가능함을 의미한다.

    To train this friction model, which is decoupled from external torque, training data was collected by driving an excitation trajectory without external contact ($\tau_{ext}=0$). Under this condition, the measured $\tau_{res}$ corresponds to the ground-truth friction that must be compensated. Particularly, the transition friction in the static-to-dynamic regime exhibits rapidly changing, high-frequency characteristics. \jh{To enhance the learning of high-frequency characteristics}, we adopted the Multi-head TCN architecture proposed in \cite{paper13_PINN}, which processes the residual by separating it into low- and high-frequency components.  
    
    This architecture employs a shared TCN backbone that processes the temporal input sequence through dilated convolutional layers. The feature vector from the final timestep is input to two separate linear heads that independently predict the low- and high-frequency components of the friction residual. \ks{We use a kernel size of $k=4$ to give the model more time window, so the initial friction state still affects the estimates during the transition phase. }%\jh{To ensure that the initial friction state during the transition phase is retained for an extended duration, we selected a large kernel size of 4.} 
    The model is trained using a weighted combination of mean squared errors from both outputs.

    % 이러한 외력과 분리된 마찰 모델을 학습하기 위해, 학습 데이터는 외부 접촉($\tau_{ext}=0$)이 없는 상태에서 excitation trajectory를 구동하며 수집되었다. 이 조건 하에서는 측정된 $\tau_{residual}$이 보상해야하는 Ground Truth frciotn에 해당한다. 특히, 보상해야 하는 정지-동적 전환 구간의 마찰(transition friction)은 매우 빠르게 변화하는 high-frequency 특성을 가지므로 효과적인 학습을 위해, \cite{paper13_PINN}에서 제안하는 잔차를 저주파와 고주파 성분으로 분리하여 처리하는 Multi-head TCN 아키텍처를 채택하였다.

    The final external torque estimation of the proposed architecture is defined as follows:
    \begin{equation}
        \hat{\tau}_{ext,comp} = \tau_{meas} - (\tau_{dyn} + \tau_{fric} + \tau_{TCN}).
        \label{eq:tcn_estimation}
    \end{equation}
    \rev{The complete torque estimation framework is presented in Fig. 3. The system operates at the motor communication rate to maintain synchronization between the nominal and TCN-based components. For robust detection, we employ a single dead-band threshold determined by the TCN estimator \eqref{eq:tcn_estimation} across all conditions. While the static estimator \eqref{eq:static_ext_def} permits a lower threshold, using a unified value prevents discontinuities during state transitions and sufficiently covers the error margins of the static-to-dynamic transition.}
    \vspace{6pt}
    % Although the static estimator in \eqref{eq:static_ext_def} requires only a very small dead-band for motor-noise rejection, the TCN-based estimator in \eqref{eq:tcn_estimation} needs a slightly larger dead-band to accommodate modeling errors. Because ad hoc threshold changes are undesirable, we adopt the TCN-based estimator as the default and apply it unchanged in static conditions as well.

    %While the external torque estimated via \eqref{eq:static_ext_def} in the static state requires only a very low deadband (sufficient for motor noise rejection), Equation \eqref{eq:tcn_estimation} necessitates an appropriate deadband to account for modeling errors. Consequently, to unify the dead-zone, the TCN-driven compensation model \eqref{eq:tcn_estimation} is utilized to estimate external torque across both static and dynamic regimes.
    % 제안하는 아키텍쳐의 최종적인 외력 추정은 다음과 같이 정의된다:
    % \begin{equation}
    %     \hat{\tau}_{ext} = \tau_{measure} - (\tau_{dyn} + \tau_{fric} + \tau_{TCN})
    %     \label{eq:tcn_estimation}
    % \end{equation}
    % 이때 정적 상태에서 \eqref{eq:static_ext_def}를 통해 추정하는 외력은 모터 노이즈 제거 수준의 매우 낮은 데드밴드를 가지만, \eqref{eq:tcn_estimation}는 적절한 모델 오차에 대한 데드밴드 필수적이다. 이로 인해 데드존 통일을 위해 정적 동적 구간 모두 TCN driven compensation model\eqref{eq:tcn_estimation}로 외력을 추정한다. 

        \begin{table}[t!]
        \centering
        \caption{Residual analysis of models in static and static-to-kinetic states}
        % 첫 번째 서브테이블
        \begin{subtable}[t]{0.475\textwidth}
            % \caption{}
            \centering
            \begin{tabular*}{\columnwidth}{@{\extracolsep{\fill}} c|c c c c}
                \hline
                \hline
                \multicolumn{5}{c}{RMSE(mA) in Static State} \\ \hline
                Model           & Joint 1  & Joint 2 & Joint 3 & Joint 4 \\ \hline
                Nominal Dynamic & 92.56   & 387.10 & 329.22 & 311.81 \\ 
                TCN Compensated & 18.32   & 37.73  & 43.89  & 27.88  \\ \hline
                \hline
            \end{tabular*}
        \end{subtable}
        % \hspace{0.05\textwidth} % 테이블 사이 간격
% 
        \begin{subtable}[t]{0.475\textwidth}
        % \caption{}
            \centering
            \begin{tabular*}{\columnwidth}{@{\extracolsep{\fill}} c|c c c c}
                \multicolumn{5}{c}{RMSE(mA) in Static-to-Kinetic State} \\ \hline
                Model           & Joint 1  & Joint 2 & Joint 3 & Joint 4 \\ \hline
                Nominal Dynamic & 125.07   & 333.60 & 349.84 & 275.28 \\ 
                TCN Compensated & 85.43   & 119.36  & 175.63  & 118.80  \\ \hline
                \hline
            \end{tabular*}
        \end{subtable}

        \begin{subtable}[t]{0.475\textwidth}
            % \caption{}
            \centering
            \begin{tabular*}{\columnwidth}{@{\extracolsep{\fill}} c|c c c c}
                \multicolumn{5}{c}{Std Dev(mA) in Static State} \\ \hline
                Model           & Joint 1  & Joint 2 & Joint 3 & Joint 4 \\ \hline
                Nominal Dynamic & 92.46   & 384.17 & 329.07 & 311.81 \\ 
                TCN Compensated & 18.26   & 37.53  & 43.44  & 27.86  \\ \hline
                \hline
            \end{tabular*}
        \end{subtable}

        \begin{subtable}[t]{0.475\textwidth}
        % \caption{}
            \centering
            \begin{tabular*}{\columnwidth}{@{\extracolsep{\fill}} c|c c c c}
                \multicolumn{5}{c}{Std Dev(mA) in Static-to-Kinetic State} \\ \hline
                Model           & Joint 1  & Joint 2 & Joint 3 & Joint 4 \\ \hline
                Nominal Dynamic & 125.08   & 329.77 & 349.70 & 275.29 \\ 
                TCN Compensated & 85.29   & 119.05  & 175.60  & 118.52  \\ \hline
            \end{tabular*}
        \end{subtable}
        \label{tab:error}
        \vspace{-6pt}
    \end{table}
    
\section{EXPERIMENT AND RESULT}
    \subsection{Error Analysis and Model Validation}
    This section quantitatively analyzes the motor torque estimation accuracy under force-free conditions for the nominal dynamic model, derived from identification-based calculation of $\tau_{dynamic}$ and $\tau_{fric}$, and the TCN compensated model, which includes the compensation term $\tau_{TCN}$. The error analysis calculates $\tau_{error}$ according to Eq. \eqref{eq:2}, assuming $\tau_{ext}=0$. As the motor used in this study does not provide an official motor torque constant $K_t$, all torque values in the experiments are replaced by current values with the unit (mA).

    Experimentally, static state data were collected when the joint velocity remained below a threshold of 0.0001 $rad/s$. % For the static-to-kinetic transition, the analysis utilized data averaging 70 ms (30 data points). This window begins when the motor exceeds the threshold velocity and concludes at the point where the residuals of both models equalize. 
    \rev{For the static-to-kinetic transition, data were collected from the moment the velocity exceeded the threshold to the point where the residuals of both models equalized. From this analysis, an average window size of 70 ms (30 data points) was determined.} The results for the RMSE and STD of the residual between the actual current value and the model are shown in Table~\ref{tab:error}. In the Static state, the compensator reduced the RMSE of the current residual from the nominal model's average of 280.17 mA to 31.96 mA, an 88.6\% reduction; this indicates that the static friction effect was, while not perfectly, substantially mitigated. In the static-to-kinetic transition, the average RMSE for all joints decreased from 270.95 mA to 124.81 mA, a 54.8\% reduction. The standard deviation of the residual also showed similar values to the RMSE, implying that the model has low bias. Fig. 4 shows the probability density function and confidence interval for both models, and this reduction in error variance is important as it enables the setting of a smaller estimation dead band, allowing the system to react more sensitively to subtle external forces.
    
    \begin{figure}[t!]
        \centering
        \includegraphics[width=3.3in]{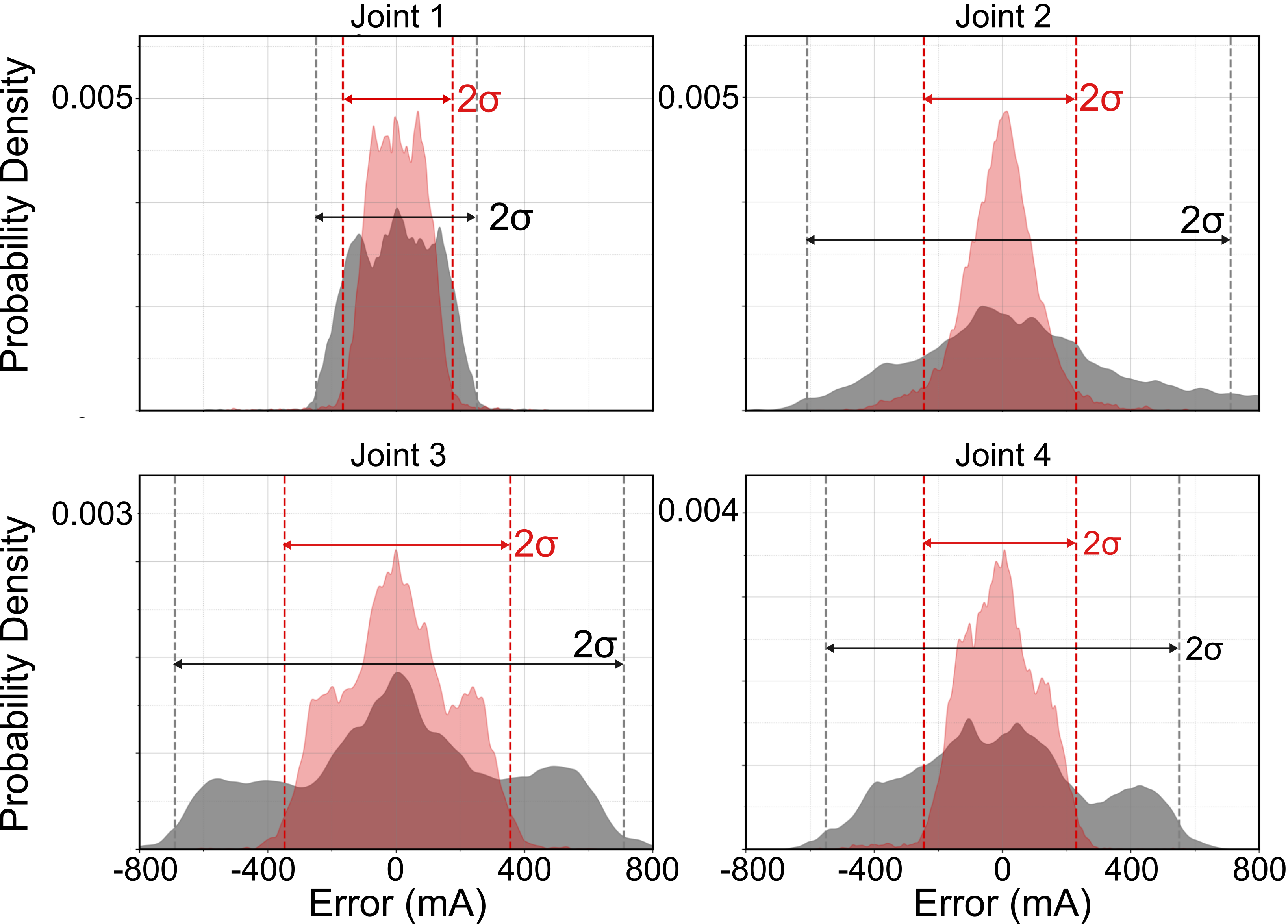}
        \caption{Joint-wise standard deviation (STD) probability density function (PDF) in the static-to-kinetic transition region. Gray: identification-based nominal model; Red: TCN-compensated model.}
        \label{figure label 0}
    \end{figure}
    \begin{figure}[t!]
        \centering
        \includegraphics[width=3.3in]{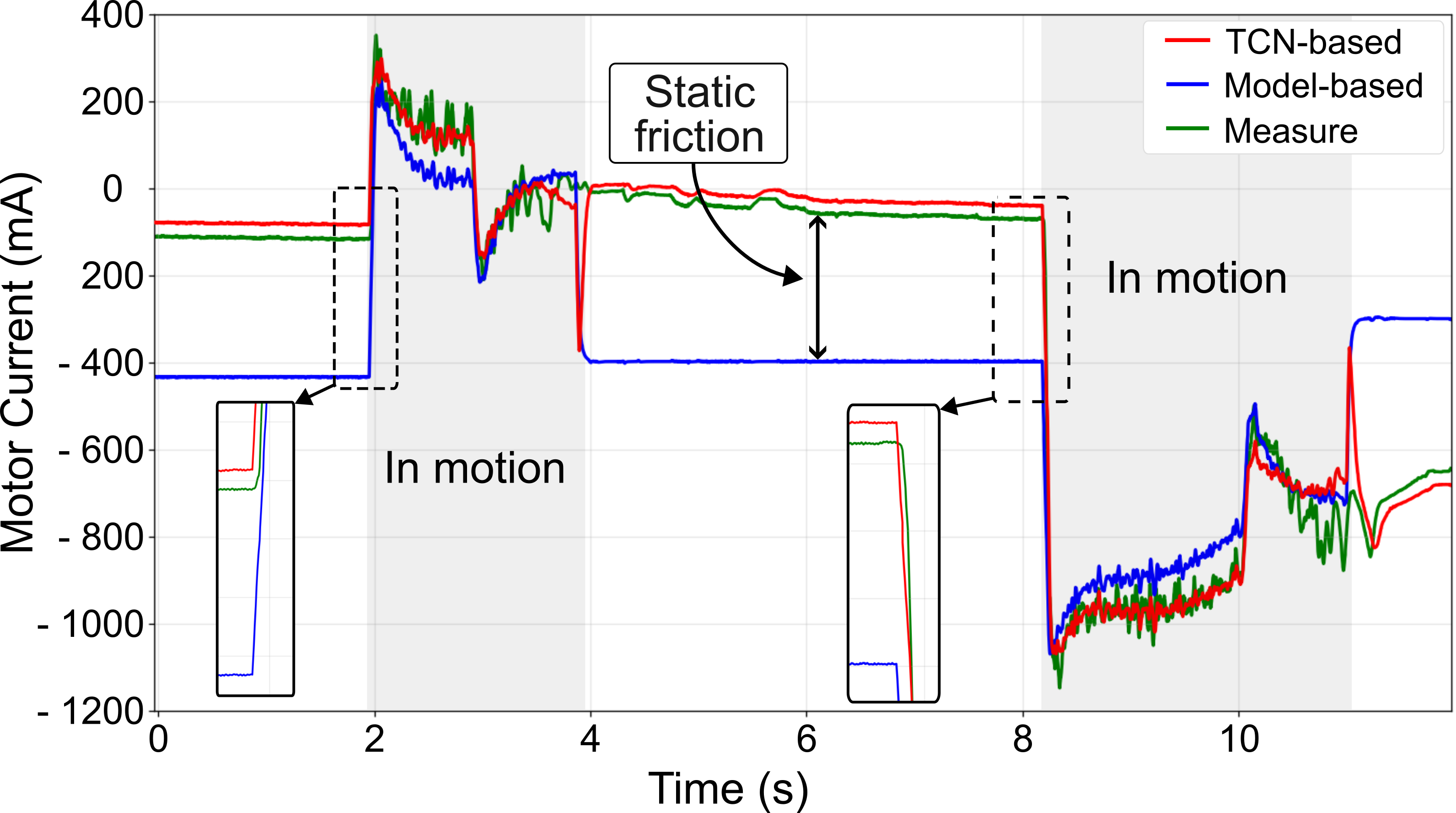}
        \caption{Model predictions for Joint 4 during static-to-kinetic transitions. \jh{Green: measured motor current corresponding to the ground truth;} Blue: identification-based model (without compensation); Red: TCN-compensated model.} %The sequence begins from a static state at time 0s}
        \vspace{-6pt}
        \label{figure label 0}
    \end{figure}

% 큰피겨
    \begin{figure*}[htbp]
        \centering
        \includegraphics[width=6.9in]{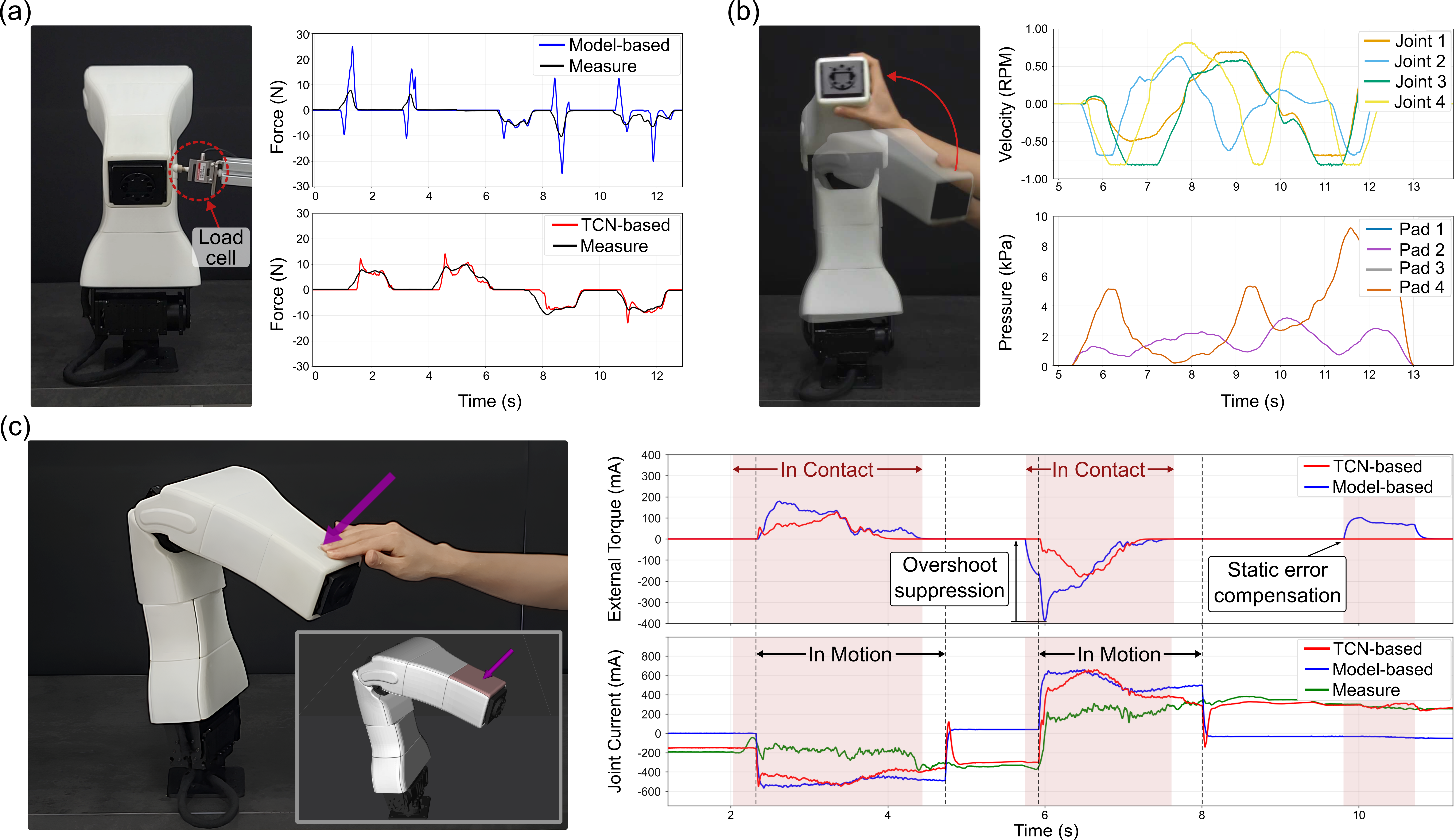}
        \jh{
        \caption{Demonstration of the TCN-based compensator performance and sensor fusion during physical human-robot interaction scenarios. 
        (a) Comparison of system response during static-to-kinetic transition without and with the TCN compensator;
        (b) Joint trajectory and pneumatic sensor data during a shear-force-dominant task compliance;
        (c) Estimated external force and motor torque for Joint 3 in a general kinesthetic teaching scenario, comparing TCN compensator and nominal model responses.
        }
        \label{figure label 06}}
    \end{figure*}

    Fig. 5 compares the actual motor current with the estimates of both models for an arbitrary motion of joint 4. \jh{Throughout the stationary period, the TCN model effectively cancels the large static-friction residual present in the nominal model. Around the motion onset and termination, the TCN model produces smooth current estimates without spike-like errors by carrying over the initial friction. However, there is a estimation error immediately after stopping because $\tau_{fric, s}=0$ during motion, the TCN's input window initially lacks stationary samples when returning to rest, causing a short mismatch until the window fills. This error is not a problem since this phase satisfies the static condition and external force estimation is still possible using Equation \eqref{eq:static_ext_def} (model free method). Additionally, performance improvements were observed during motion since the joint's kinetic state is input to the TCN; however, as this is outside the scope of this study, so we omit further analysis.}

\subsection{Demonstration}
    Effective kinesthetic teaching demands several key system properties for intuitive pHRI. The system must exhibit low force thresholds and low damping characteristics, enabling the operator to initiate motion with minimal force and sustain teaching without fatigue \cite{paper28_2022}. Furthermore, the system response must be predictable and consistent, accurately reflecting user intent without overshoot or undershoot.
    These properties, however, are difficult to achieve using a single sensing modality.
    
    Proprioceptive-based approaches inherently face a trade-off between responsiveness and stability that a dead band set for stability results in poor responsiveness, while lowering damping induces system instability \cite{paper29_2024}. 
    
    Conversely, approaches relying solely on tactile data are limited in providing sufficient information for precise task compliance. Although tactile signals are adept at resolving the location and approximate magnitude of discrete contact points, they often fail to adequately account for complex multi-point contact scenarios, such as grasping, where the individual forces distributed across the robot’s surface may not align with the net force vector required for the task.
    
    Here we propose proprioceptive-tactile fusion framework that addresses these limitations. The system achieves robust task compliance unattainable by tactile-only sensing, while simultaneously mitigating the responsiveness limitations inherent to proprioceptive methods, thereby providing an intuitive and stable user experience. For validation, we performed a kinesthetic teaching demonstration using admittance control (Fig. 6).

    \jh{To ensure high responsiveness, we configured a small dead band of 1.5$\sigma$ (∼86.6\%) for all demonstration scenarios, with $\sigma$ being the standard deviation of the estimation error measured during the static-to-kinetic transition (Table I). Although such a low threshold is generally undesirable for safe pHRI, our robotic skin directly senses contact, providing reliable contact-state estimation and preserving both stability and responsiveness.}

    \jh{The first demonstration compares the force estimation performance of the TCN compensator relative to a nominal model (Fig. 6(a)). The robot executes compliance motion according to the external forces vector, with forces applied using a load cell. To evaluate external force estimation performance during static-to-kinetic transitions, weak and slow forces were applied in short, repeated intervals. Without compensation, applying a force to the robot resulted in significant overshoot and undershoot. In cases where the previous and current force directions were identical, these effects were particularly severe. Incorporating the TCN model eliminated undershoot and substantially reduced overshoot.}

    \jh{To evaluate task compliance during kinesthetic teaching, we guided the robot by firmly grasping the end-effector and executing complex trajectories as presented in Fig. 6(b). In this setting, pneumatic-pad signals vary with contact area and shape, so tactile measurements alone are unreliable for characterizing the interaction. Instead, the proposed sensor-fusion framework estimates the contact location from the pads and maps the resulting joint-level torque to a task-compliance trajectory.}

    \jh{While the previous demo evaluated task compliance in a grasp, the next presents a more general kinesthetic teaching with non-prehensile interactions (Fig. 6(c)). This scenario consists of three sequential contacts. In the second contact, the TCN compensator addresses two issues. First, it removes the static-friction error at contact onset. This error can appear even before motion begins because the dead band is small. Second, it reduces the overshoot that occurs when the motor starts responding to the external force. During the third contact, the compensator again eliminates the static-friction error. In this demonstration, the executed motion follows the compliance trajectory generated by the TCN model, and the nominal model is evaluated by computing its prediction along this TCN-generated motion.}

\section{DISCUSSION}
    % 이 논문에서, 우리는 대면적 로봇 피부의 접촉 정보와 motor current 정보 통하여, Tactile 센서만으로는 얻지 못하는 전단방향 힘 성분을 추정을 가능하게하고, proprioceptive방식의 높은 데드밴드 문제를 완화할 수 있는 새로운 sensor fusion 방법론을 제안하였다. 기본 nominal dynamic, friction model로는 설명하지 못하던 정지마찰을 tactile 센서의 접촉 정보를 통해 추정하였으며, 동적 전환 구간에서의 예측 불연속성으로 인한 오차들은 새로운 입력 구조를 가지는 TCN을 통하여 보상하였다. 우리는 실험을 통해 TCN모델의 데드밴드를 nominal model에 비해 53.86\% 낮게 설정할 수 있음을 보였으며, 실제 교시상황 데모를 통해 TCN comepensator가 교시상황에서 오버슈트 및 언더슈트 보상이 적절히 이루어지고 tactile센서와 motor current sensor fusion이 높은 시너지를 보임을 확인하였다.
    
    % 우리의 TCN이 motion onset 시점은 효과적으로 보상하지만, Dynamic 상태에서 Static (with contct) 상태로 전환되는 과정에서는 한계를 보입니다. 이 상태 천이 구간에서는 운동 직후의 예측 불가능한 마찰 상태로 인해, 유효한 $\hat{\tau}_{fric, s}$을 확보하기 어렵다. 전체 시스템 아키텍처가 준수 모드(compliance mode)를 강제로 종료하도록 보장하기 때문에 이 현상이 critical stability concern를 야기하지는 않는다. 그럼에도 불구하고, 이 kinetic-to-static 전환 과정을 정확하게 모델링하는 것에 대한 후속 연구가 필요하다.
    % 덧붙여, 본 연구에서는 크게 강조되지 않았던 Equation $\eqref{eq:static_ext_def}$의 독립적인 잠재력을 강조할 필요가 있다. 이 방정식은 동역학 모델 없이 정지 상태에서 $\hat{\tau}_{ext}$를 추정하며, 이는 오직 모터 노이즈 제거 수준의 '최소한의 데드존(minimal deadband)'만을 요구합니다. 이러한 특성은 향후 로봇 암 자체를 고정밀 햅틱 유저 인터페이스(high-fidelity haptic user interface) 또는 정적 토크 입력 디바이스로 활용하는 연구로 확장할 때 강력한 잠재력을 제공할 것이다.

    In this paper, we present a sensor-fusion method that combines contact information from large-area robotic skin with motor-current data. By exploiting tactile cues, we estimate a static-friction baseline in a model-free manner. We use this baseline (static friction), together with kinematic states, as direct inputs to a temporal convolutional network (TCN) that compensates friction at motion onset. We found that the TCN model reduced the dead-band by 53.86\% relative to the uncompensated baseline, thereby improving contact sensitivity. The TCN also mitigated overshoot and undershoot appropriately, suggesting the benefits of the sensor fusion in realizing natural and smooth interaction.
    %fusing tactile cues with motor-current sensing.
    
    %Through our method, we demonstrated that the dead band for kinesthetic teaching can be reduced

    %Static friction, which nominal dynamic and friction models fail to explain, was estimated using tactile contact information. Furthermore, errors resulting from prediction discontinuities during dynamic transition phases were compensated via a TCN featuring a novel input structure. Through experiments, we demonstrated that the dead band of the TCN model could be reduced by 53.86\% compared to the nominal model. Additionally, a practical demonstration confirmed that the TCN compensator appropriately addresses overshoot and undershoot during teaching scenarios, verifying the significant synergy achieved by fusing tactile and motor current data.
    
    While the TCN mitigates errors at motion onset, it is less reliable during the transition from dynamic motion to static state. Immediately after motion ceases, the friction state is unpredictable, making the estimate of static friction torque $\hat{\tau}_{fric, s}$ unreliable. Nonetheless, this does not compromise stability, because the compliance mode is terminated as the contact disappears. Future work will model this issue more thoroughly.
    %Although our TCN effectively compensates for motion onset, it exhibits limitations during the transition from a Dynamic state to a Static (with contact) state. In this state transition phase, obtaining a valid $\hat{\tau}_{fric, s}$ is challenging owing to the unpredictable friction state immediately following motion. This phenomenon does not pose a critical stability concern, as the overall system architecture ensures the termination of compliance mode via tactile data. Nonetheless, follow-up research is required to accurately model this kinetic-to-static transition process.
    
    % Haptic interface는 예전부터 highly backdrivable한 구동 메커니즘을 사용해왔기 때문에, 우리 연구가 딱히 큰 이득이 안됨 아하 넵
    %Furthermore, it is necessary to highlight the independent potential of Equation $\eqref{eq:static_ext_def}$, which was not heavily emphasized in the present study. This equation determines $\hat{\tau}_{ext}$ in a static state without requiring a dynamics model, demanding only a minimal dead band equivalent to the level of motor noise cancellation. This characteristic offers significant potential for future work, enabling the extension of this research toward utilizing the robot arm itself as a high-fidelity haptic user interface or a static torque input device.

    \section*{ACKNOWLEDGMENT}
    This work was supported in part by the National Research Foundation of Korea (NRF) grant funded by the Korea government (MSIT) (RS-2024-00352818), in part by the NRF grant funded by the MSIT (RS-2025-25448259), in part by Basic Science Research Program through the NRF funded by the Ministry of Education (RS-2025-25420118), and in part by the Institute of Information \& Communications Technology Planning \& Evaluation (IITP) grant funded by the Korea government (MSIT) (RS-2025-25442149, LG AI STAR Talent Development Program for Leading Large-Scale Generative AI Models in the Physical AI Domain).
    % This work was supported by the National Research Foundation of Korea(NRF) grant funded by the Korea government(MSIT)(RS-2024-00352818)

    % Reference (bibliography)
    \bibliographystyle{IEEEtran}
    \bibliography{reference.bib}
    
\end{document}